\documentclass[letterpaper, 10 pt, conference]{ieeeconf}

\usepackage{graphicx}
\usepackage{amsmath}
\usepackage{amsfonts}
\usepackage{amssymb}
\usepackage{algorithm}
\usepackage{algorithmic}
\usepackage{url}

\IEEEoverridecommandlockouts
\newcommand{\cM}{\mathcal{M}}

\newcommand{\cS}{\mathcal{S}}
\newcommand{\cA}{\mathcal{A}}
\newcommand{\cT}{\mathcal{T}}
\newcommand{\cR}{\mathcal{R}}

\newcommand{\bR}{\mathbb{R}}
\newcommand{\bE}{\mathbb{E}}

\newcommand{\Vpi}{V^{\pi}}
\newcommand{\Qpi}{Q^{\pi}}
\newcommand{\Api}{A^{\pi}}
\newcommand{\pith}{\pi_\theta}

\title{\LARGE \bf
Reinforcement learning for non-prehensile manipulation: \\ Transfer from simulation to physical system
}

\author{Kendall Lowrey$^{1,2}$, Svetoslav Kolev$^{1}$, Jeremy Dao$^{1}$, Aravind Rajeswaran$^{1}$ and Emanuel Todorov$^{1,2}$
\thanks{*This work was supported by the NSF.}
\thanks{$^{1}$ University of Washington, $^{2}$ Roboti LLC.}%
\thanks{Correspond to: \tt{klowrey@cs.washington.edu}}
}

\begin{document}

\maketitle
\thispagestyle{empty}
\pagestyle{empty}

\begin{abstract}

Reinforcement learning has emerged as a promising methodology for training robot controllers. However, most results have been limited to simulation due to the need for a large number of samples and the lack of automated-yet-safe data collection methods. Model-based reinforcement learning methods provide an avenue to circumvent these challenges, but the traditional concern has been the mismatch between the simulator and the real world. Here, we show that control policies learned in simulation can successfully transfer to a physical system, composed of three Phantom robots pushing an object to various desired target positions. We use a modified form of the natural policy gradient algorithm for learning, applied to a carefully identified simulation model. The resulting policies, trained entirely in simulation, work well on the physical system without additional training. In addition, we show that training with an ensemble of models makes the learned policies more robust to modeling errors, thus compensating for difficulties in system identification. The results are illustrated in the accompanying video.
\end{abstract}

\section{INTRODUCTION}

Non-prehensile object manipulation remains a challenging control problem in robotics. In this work, we focus on a particularly challenging system using three Phantom robots as fingers. These are haptic robots that are torque-controlled and have higher bandwidth than the fingers of existing robotic hands. In terms of speed and compliance (but not strength), they are close to the capabilities of the human hand. This makes them harder to control, especially in non-prehensile manipulation tasks where the specifics of each contact event and the balance of contact forces exerted on the object are very important and need to be considered by the controller in some form.

Here we develop a solution using Reinforcement Learning (RL). We use the MuJoCo physics simulator~\cite{todorov2012mujoco} as the modeling platform and fit the parameters of the simulation model to match the real system as closely as possible. For policy learning with the model, we use a normalized natural policy gradient (NPG) method \cite{Kakade01,Peters,Rajeswaran17generalization}. While Reinforcement Learning (RL) methods such as NPG are in principle model-free, in practice they require large amounts of data. In the absence of an automatic way to generate safe exploration controllers, learning is largely possible only in simulation. Indeed the vast majority of recent results in continuous RL have been obtained in simulation. These studies often propose to extend the corresponding methods to physical systems in future work, but the scarcity of such results indicates that `sim-to-real' transfer is harder than it seems. The few successful applications to real robots have been in tasks involving position or velocity control that avoid some difficulties of torque-controlled platforms.

To obtain an accurate simulation model, we fit the parameters of the simulation model using a physically consistent system identification procedure~\cite{kolev2015physically} developed specifically for identification in contact rich settings such as the one we study. While true model-free RL may one day become feasible, we believe leveraging the capabilities of a physics simulator will always help speed up the learning process.

As with any controller developed in simulation, performance on the real system is likely to degrade due to modeling errors. To assess, as well as mitigate, the effects of these errors, we compare learning with respect to three different models: (i) the nominal model obtained from system identification; (ii) a modified model where we intentionally mis-specify some model parameters; (iii)~an ensemble of models where the mean is wrong but the variance is large enough to include the nominal model. The purpose of (ii) is to simulate a scenario where system identification has not been performed well, and we wish to study the performance degradation. The goal with ensemble approaches~\cite{mordatch2015ensemble, rajeswaran2016epopt,Peng2017SimtoReal} is to study if including a distribution over models can compensate for inaccuracies in system identification during control tasks. We find that (i) achieves the best performance as expected, and (iii) is robust but suffers a degradation in performance. Videos of the trained policies is available at: \url{https://sites.google.com/view/phantomsim2real} 

\subsection{Related Work}

There are many methods towards developing safe and robust robot controllers. Robot actions that involve dynamic motions require not only precise control execution, but also robust compensation when the action inevitably does not go according to plan--the physics of the real world are notoriously uncooperative. Control methods that depend on physical models, whether or not with reduced or simplified models, are able to produce dynamic actions \cite{tassa2012synthesis,mordatch2012discovery,tassa2014control,feng2015optimization,koenemann2015whole,raibert2008bigdog}. They frequently rely on physics simulations for testing purposes, before usage on real hardware. This step is critical, as any modeling errors can significantly contribute to poor performance or even hardware damage \cite{nguyen2011model}. Including uncertainty in the planning stage is one way to avoid this problem, and may also enable model learning simultaneously \cite{mordatch2016combining,pan2015data,wang2010optimizing,lee2017gp,ross2012agnostic}. These model centric approaches offer strong performance expectations, but unless uncertainty or robustness is explicitly taken into account, may be brittle to external unknowns \cite{johnson2015team}.

On the other hand, Reinforcement Learning offers a means to directly learn from the robot's experience \cite{peters2003reinforcement,kaelbling1996reinforcement}. The difficulty, of course, is where the robot's experiences come from: as RL algorithms may need significant amounts of data, doing this on hardware may be infeasible \cite{kober2013reinforcement}. Directly training on hardware has been feasible in some cases \cite{gu2017deep,deisenroth2011pilco}, but domains with many dimensions or highly nonlinear dynamics will always require more data to sufficiently explore, human demonstrations for imitation learning, and/or parameterized explorations \cite{chebotar2017path,yahya2016collective,kober2009policy,Rajeswaran2017dexterous}. Another common issue with learning in the real world is how to reset the state of the system, with some work being done \cite{montgomery2017reset}. For sensitive and delicate systems, the only safe place to perform learning is in simulation. Transferring to real hardware can take many approaches as well, either through adaptation \cite{menache2005basis,barrett2010transfer}, or incorporating uncertainty \cite{cutler2014reinforcement,abbeel2006using}.

This work focuses on using a physics simulator to train policies for manipulation using reinforcement learning. As the manipulator is non-prehensile, we do not use any demonstrations or guide the policy search. To facilitate transfer to hardware, we also avoid the use of an estimator (i.e. the use of a model to predict state like a Kalman filter) by learning a function that directly converts from sensor values to motor torques. The policy is then transfered to the hardware for evaluation, and show that even for incorrect models used during training, useful policies are obtained by using an ensemble of models. Sections 2 and 3 detail the RL problem formulation and solution. Section 4 explains the hardware platform and details of the manipulation task are in section 5. Finally Section 6 contains the results and Section 7 the discussion.

\section{PROBLEM FORMULATION}

We model the control problem as a Markov decision process (MDP) in the episodic average reward setting, which is defined using the tuple: $\cM = \lbrace \cS, \cA, \cR, \cT, \rho_0, T \rbrace$. 
$\cS \subseteq \bR^n$, $\cA \subseteq \bR^m$, and \hbox{$\cR: \cS \times \cA \rightarrow \bR$} are (continuous) set of states, set of actions, and the reward function respectively. $\cT: \cS \times \cA \rightarrow \cS$ is the stochastic transition function; $\rho_0$ is the probability distribution over initial states; and $T$ is the maximum episode length. We wish to solve for a stochastic policy of the form $\pi: \cS \times \cA \rightarrow \bR$, which optimizes the average reward accumulated over the episode. Formally, the performance of a policy is evaluated according to: 
\begin{equation}
\eta(\pi) = \frac{1}{T} \ \bE_{\pi, \cM} \Bigg[ \sum_{t=1}^T r_t \Bigg].
\end{equation}
In this finite horizon rollout setting, we define the value, $Q$, and advantage functions as follows:
$$ \Vpi(s,t) = \bE_{\pi, \cM} \Bigg[ \sum_{t'=t}^{T} r_{t'} \Bigg] $$ 
$$ \Qpi(s,a,t) = \bE_{\cM} \Big[\cR(s,a) \Big] + \bE_{s'\sim \cT(s,a)} \Big[ \Vpi(s', t+1) \Big] $$
$$ \Api(s,a,t) = \Qpi(s,a,t) - \Vpi(s,t) $$
We consider parametrized policies $\pith$, and hence wish to optimize for the parameters $(\theta)$. In this work, we represent $\pith$ as a multivariate Gaussian with diagonal covariance. In our experiments, we use an affine policy as our function approximator, visualized in figure \ref{fig:heatmap}.

\begin{algorithm}[h]
	\caption{Distributed Natural Policy Gradient}
	\label{alg:DNPG}
	\begin{algorithmic}[1]
		\STATE Initialize policy parameters to $\theta_0$
		\FOR{$k=1$ {\bfseries to} $K$}
		\STATE Distribute Policy and Value function parameters.
		\FOR{$w=1$ {\bfseries to} $N_{workers}$}
		\STATE Collect trajectories $\lbrace \tau^{(1)}, \ldots \tau^{(N)} \rbrace$ by rolling out the stochastic policy $\pi(\cdot;\theta_k)$.
		\STATE Compute $\nabla_\theta \log \pi(a_t | s_t ; \theta_k)$ for each $(s,a)$ pair along trajectories sampled in iteration $k$.
		\STATE Compute advantages $A^\pi_k$ based on trajectories in iteration $k$ and approximate value function $V^\pi_{k-1}$.
		\STATE Compute policy gradient according to eq. (2).
		\STATE Compute the Fisher matrix (4).
		\STATE Return Fisher Matrix, gradient, and value function parameters to central server.
		\ENDFOR
		\STATE Average Fisher Matrix gradient, and perform gradient ascent (5)
		\STATE Update parameters of value function.
		\ENDFOR
	\end{algorithmic}
\end{algorithm}

\section{Method}

\subsection{Natural Policy Gradient}

Policy gradient algorithms are a class of RL methods where the parameters of the policy are directly optimized typically using gradient based methods. Using the score function gradient estimator, the sample based estimate of the policy gradient can be derived to be:~\cite{Williams92}:
\begin{equation}
\hat{g} = \frac{1}{NT} \ \sum_{i=1}^{N} \sum_{t=1}^T \nabla_\theta \log \pith(a^i_t|s^i_t) \hat{\Api}(s^i_t, a^i_t, t)
\end{equation}
A straightforward gradient ascent using the above gradient estimate is the REINFORCE algorithm~\cite{Williams92}. Gradient ascent with this direction is sub-optimal since it is not the steepest ascent direction in the metric of the parameter space \cite{Amari98}. Consequently, a local search approach that moves along the steepest ascent direction was proposed by Kakade~\cite{Kakade01} called the natural policy gradient. This has been expanded upon in subsequent works~\cite{Peters,Rajeswaran17generalization,Peters07,Schulman15,schulman2017proximal}, and forms a critical component in state of the art RL algorithms. Natural policy gradient is obtained by solving the following local optimization problem around iterate $\theta_k$:
\begin{equation}
\begin{aligned}
& \underset{\theta}{\text{maximize}}
& & g^T (\theta - \theta_k) \\
& \text{subject to}
& & (\theta - \theta_k)^T F_{\theta_k} (\theta - \theta_k) \leq \delta,
\end{aligned}
\end{equation}
where $F_{\theta_k}$ is the Fisher Information Metric at the current iterate $\theta_k$. We apply a normalized gradient ascent procedure, which has been shown to further stabilize the training process~\cite{Peters07,Schulman15,Rajeswaran17generalization}. This results in the following update rule:
\begin{equation}
\theta_{k+1} = \theta_k + \sqrt{\frac{\delta}{g^T F_{\theta_k}^{-1}g }} \ F_{\theta_k}^{-1}g.
\end{equation}
The version of natural policy gradient outlined above was chosen for simplicity and ease of implementation. The natural gradient performs covariant updates by rescaling the parameter updates according to curvature information present in the Fisher matrix, thus behaving almost like a second order optimization method. Furthermore, due to the normalized gradient procedure, the gradient information is insensitive to linear rescaling of the reward function, improving training stability. For estimating the advantage function, we use the GAE procedure~\cite{Schulman16} and use a quadratic function approximator with all the terms in $s$ for the baseline.

\subsection{Distributed Processing}

As the natural policy gradient algorithm is an on-policy method, all data is collected from the current policy. However, the NPG algorithm allows for the rollouts and most computation to be performed independently as only the gradient and the Fisher matrix need to be synced. Independent processes can compute the gradient and Fisher matrix, with a centralized server averaging these values and performing the matrix inversion and gradient step as in equation (4). The new policy is then shared with each worker. The total size of messages passed is proportional to the size of the Fisher Matrix used for the policy, and linear in the number of worker nodes. Policies with many parameters may experience message passing overhead, but the trade-off is that each worker can perform as many rollouts during sample collection without changing the message size, encouraging more data gathering (which large policies require).

A summary of the distributed algorithm we used is show in Algorithm 1.

\begin{figure}[h]
	\centering
 	\includegraphics[width=0.45\textwidth]{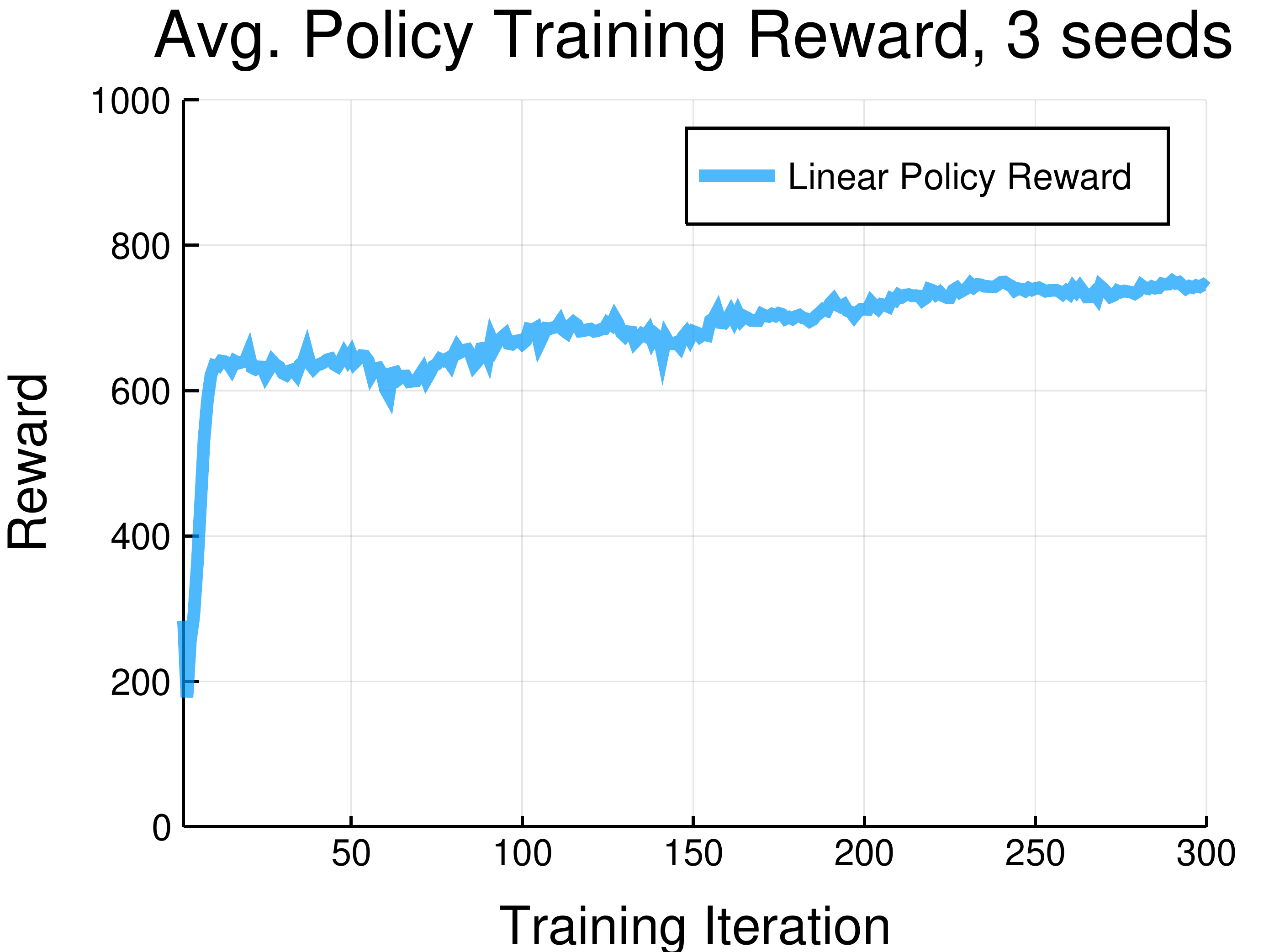}
	\caption{Learning curve of our linear/affine policy. We show here the curve for the policy trained with the correct mass as a representative curve.}
\end{figure}

We implemented our RL code and interfaced with the MuJoCo simulator with the Julia programming language \cite{bezanson2012julia}. The built-in multi-processing and multi-node capabilities of Julia facilitated this distributed algorithm's performance; we are able to train a linear policy on this task in less than 3 minutes on a 4 node cluster with Intel i7-3930k processors.

\section{HARDWARE AND PHYSICS SIMULATION}
\begin{figure}[h]
	\centering
	\includegraphics[width=0.4\textwidth]{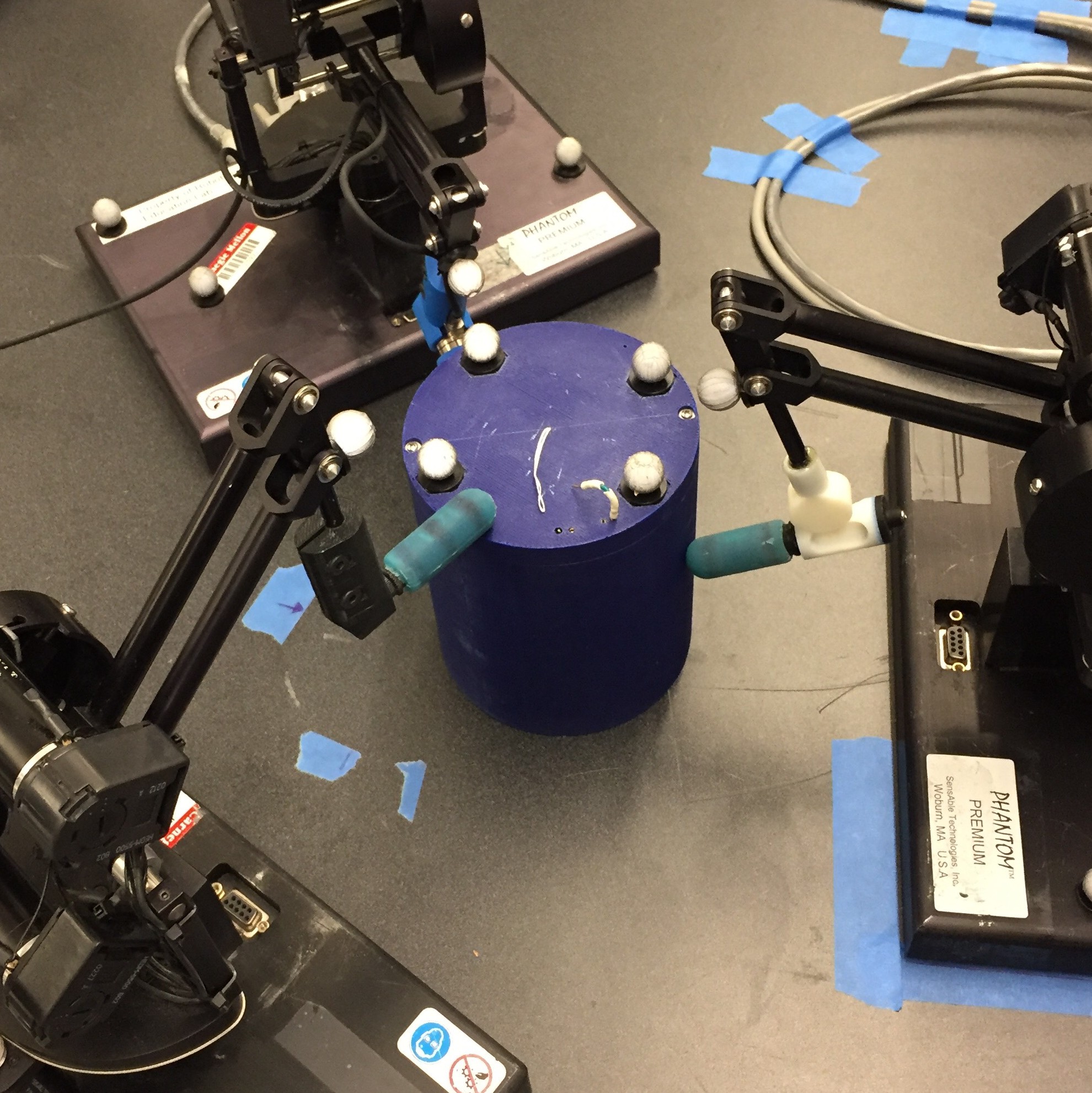}
    \caption{Phantom Manipulation Platform.}
        \vspace*{-5mm}
\label{fig:phantoms}
\end{figure}
\subsection{System Overview}

We use our Phantom Manipulation Platform as our hardware testbed. It consists of three Phantom Haptic Devices, each acting as a robotic finger. Each haptic device is a 3-DOF cable driven system shown in figure \ref{fig:phantoms}. The actuation is done with Maxon motors (Model RE 25 \#118743), three per Phantom. Despite the low gear reduction ratio, they are able to achieve 8.5N instantaneous force and 0.6N continuous force at the middle of their range of motion, with very low friction for the entire range of motion.

The three robots are coupled together to act as one manipulator. Each robot's end effector was equipped with a silicon covered fingertip to enable friction and reliable grasping of objects. The softness of the silicon coating was an additional challenge in both contact modeling and robust policy learning. For this work, we had the robots manipulating a 3D printed cylinder with a height of 14cm and diameter of 11cm, and a mass of 0.34kg.

The soft contacts, combined with the direct torque control and high power-to-weight ratio--leading to high acceleration--make this platform particularly difficult to control. Systems with more mass and natural damping in their joints naturally move more slowly and smoothly; this is not the case here. Being able to operate in this space, however, allows for the potential for high performance, dynamic manipulation, and the benefits that come with torque based control. However, this requires that we operate our robot controller at 2kHz to successfully close the loop.

\subsection{Sensing}

As we wish to learn control policies that map from observations to controls, the choice of observations are critical to successful learning. Each Phantom is equipped with 3 optical encoders at a resolution of 5K steps per radian. We use a low-pass filter to compute the joint velocities. We also rely on a Vicon motion capture system, which gives us position data at 240Hz for the object we are manipulating--we assume the object remains upright and do not include orientation. While being quite precise ($0.1mm$ error), the overall accuracy is significantly worse ($<1mm$) due, in part, to imperfect object models and camera calibrations. While the Phantoms' joint position sensors are noiseless, they often have small biases due to imperfect calibration. One Phantom robot is equipped with an ATI Nano17 3-axis force/torque sensor. This data is not used during training or in any learned controller, but used as a means of hardware / simulation comparison described in a later section. The entire system is simulated for policy training in the MuJoCo physics engine \cite{todorov2012mujoco}.

In total, our control policy has an observational input of 36 dimensions with 9 actuator outputs-- the 9 positions and 9 velocities of the Phantoms are converted to 15 positions and 15 velocities for modeling purposes due to the parallel linkages. We additionally use 3 positions for both the manipulated object and the tracked goal, with 9 outputs for the 3 actuators per Phantom. Velocity observations are not used for the object as this would require state estimation that we have deliberately avoided.

\subsection{System Identification}

System identification of model parameters was performed in our prior work \cite{kolev2015physically}, but modeling errors are difficult to eliminate completely. For system ID we collected various behaviors with the robots, ranging from effector motion in free space to infer intrinsic robot parameters, to manipulation examples such as touching, pushing and sliding between the end effector and the object to infer contact parameters. The resulting data is fed into the joint system ID and state estimation optimization procedure. As explained in \cite{kolev2015physically}, state estimation is needed when doing system ID in order to eliminate the small amounts of noise and biases that our sensors produce.

\begin{figure*}[t]
	\centering
    \vspace*{3mm}
	\includegraphics[width=0.32\textwidth]{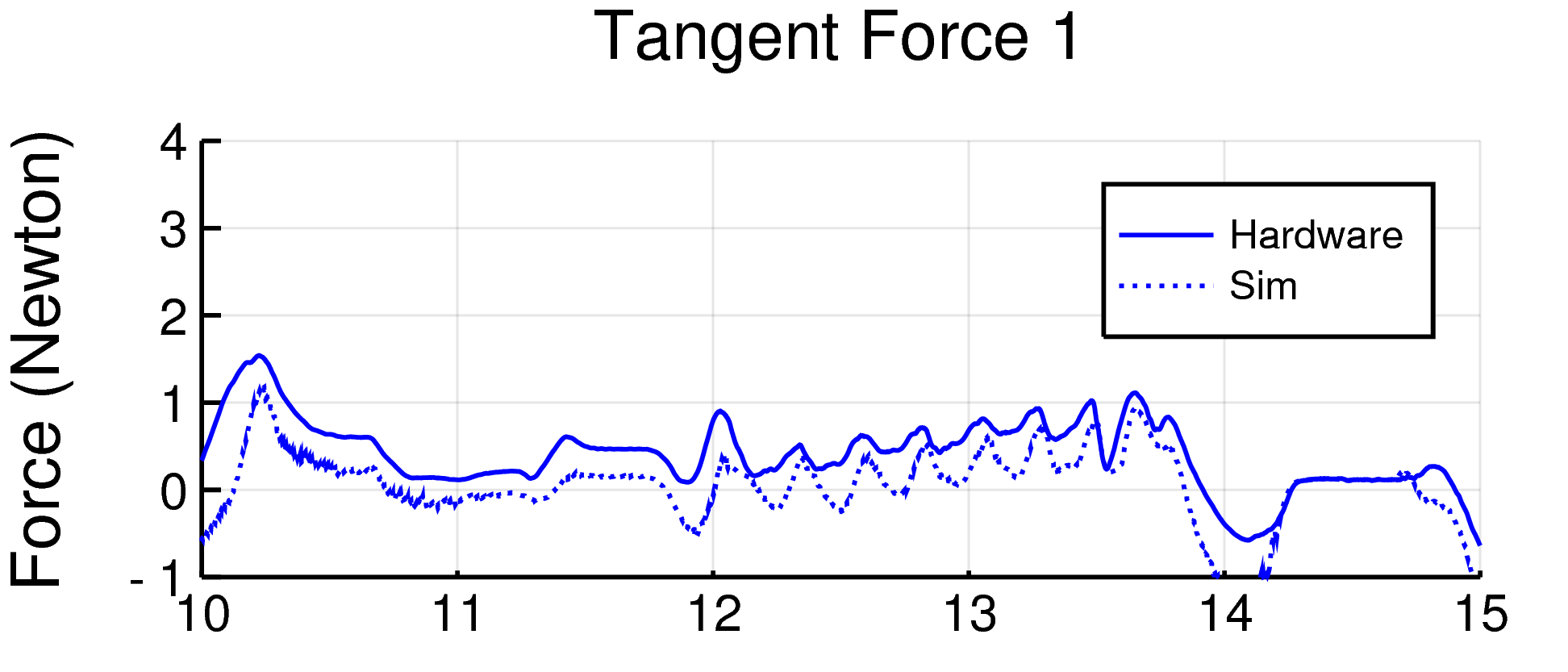}
	\includegraphics[width=0.32\textwidth]{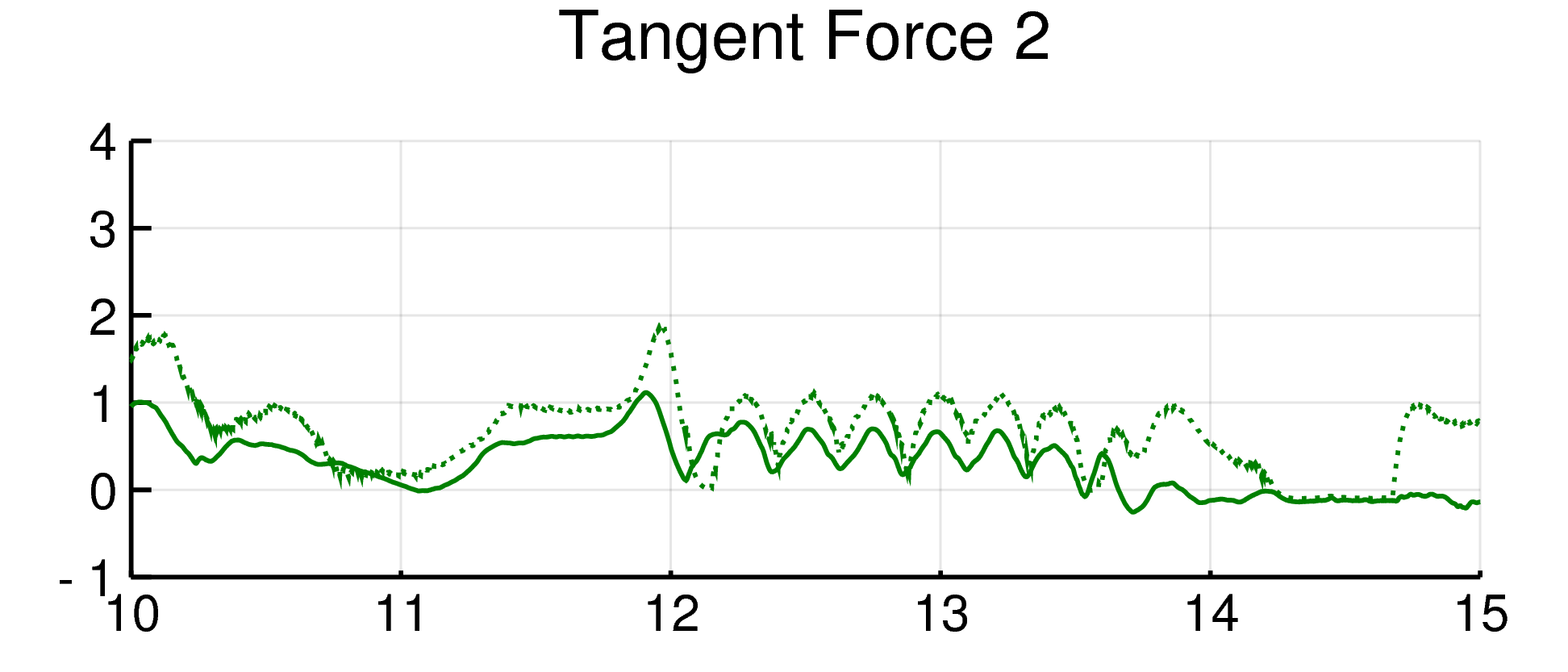}
	\includegraphics[width=0.32\textwidth]{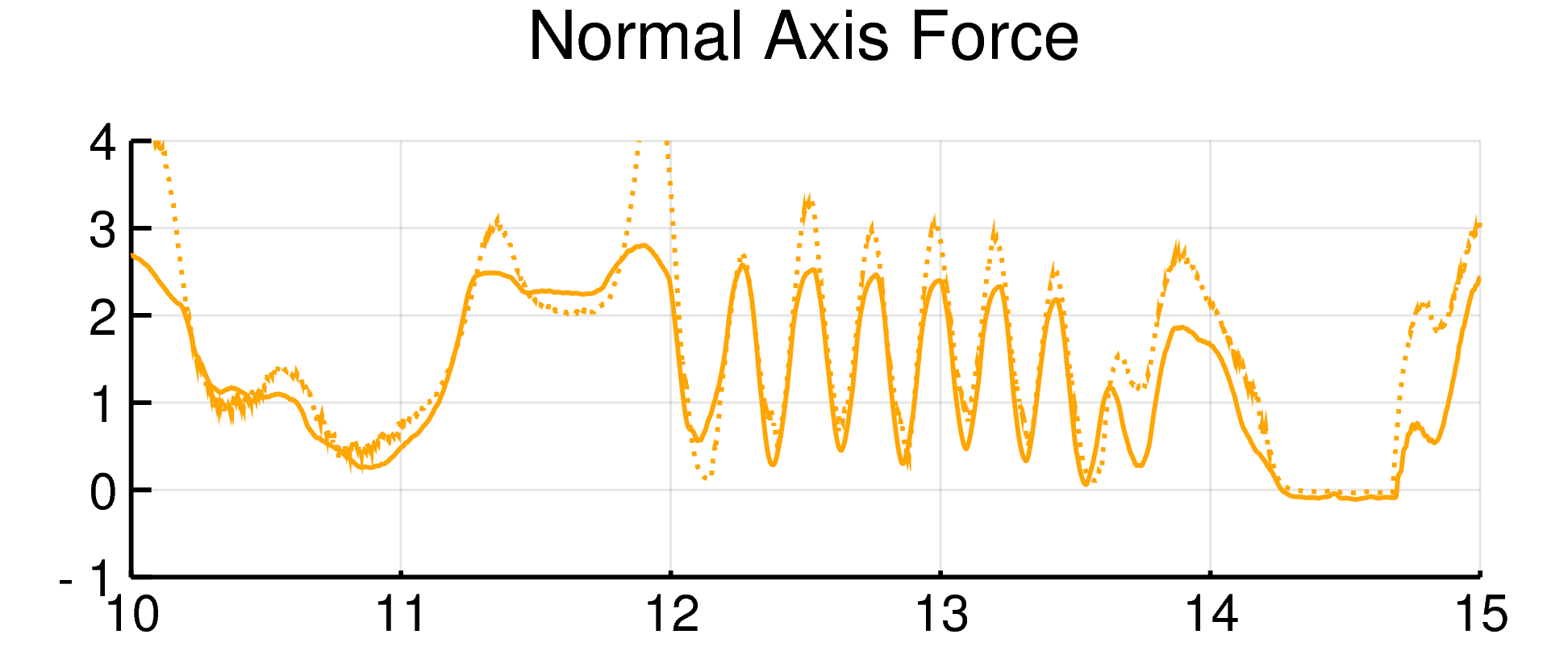}
    \includegraphics[width=0.32\textwidth]{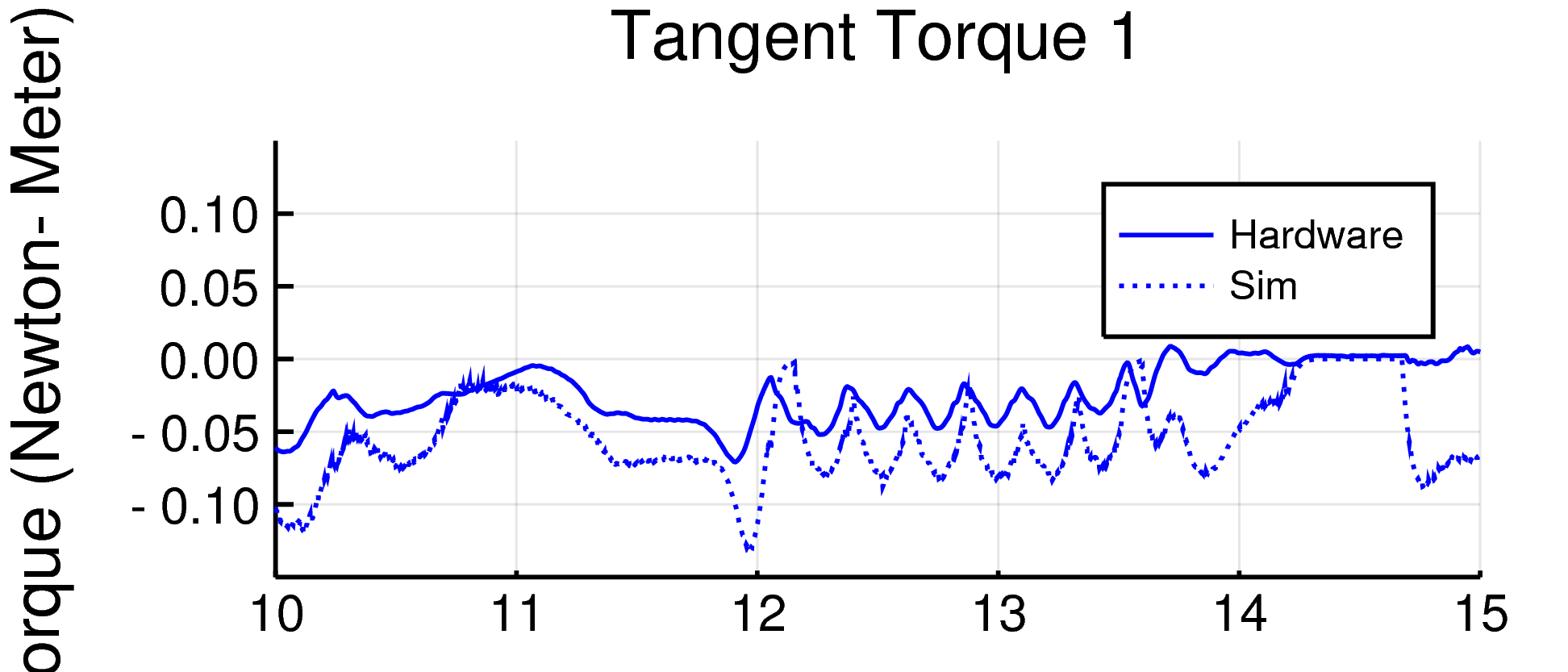}
	\includegraphics[width=0.32\textwidth]{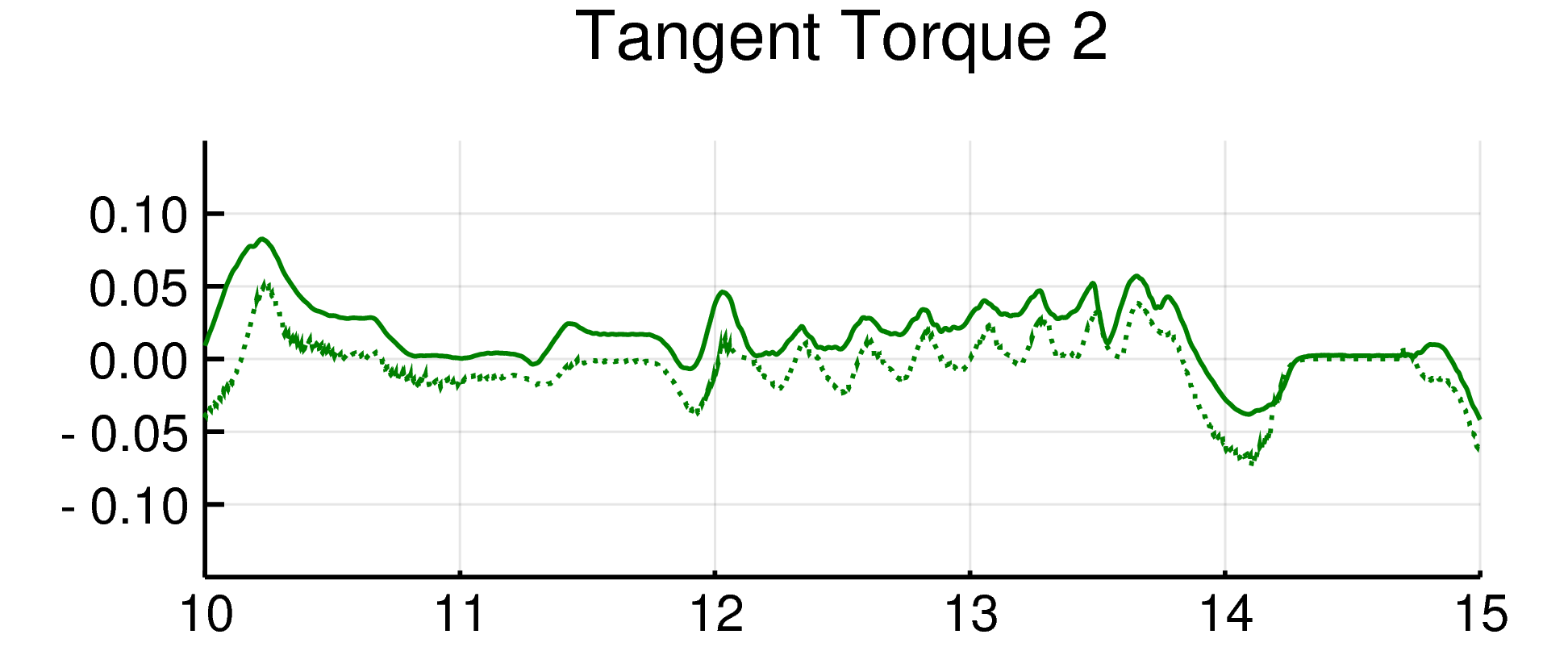}
	\includegraphics[width=0.32\textwidth]{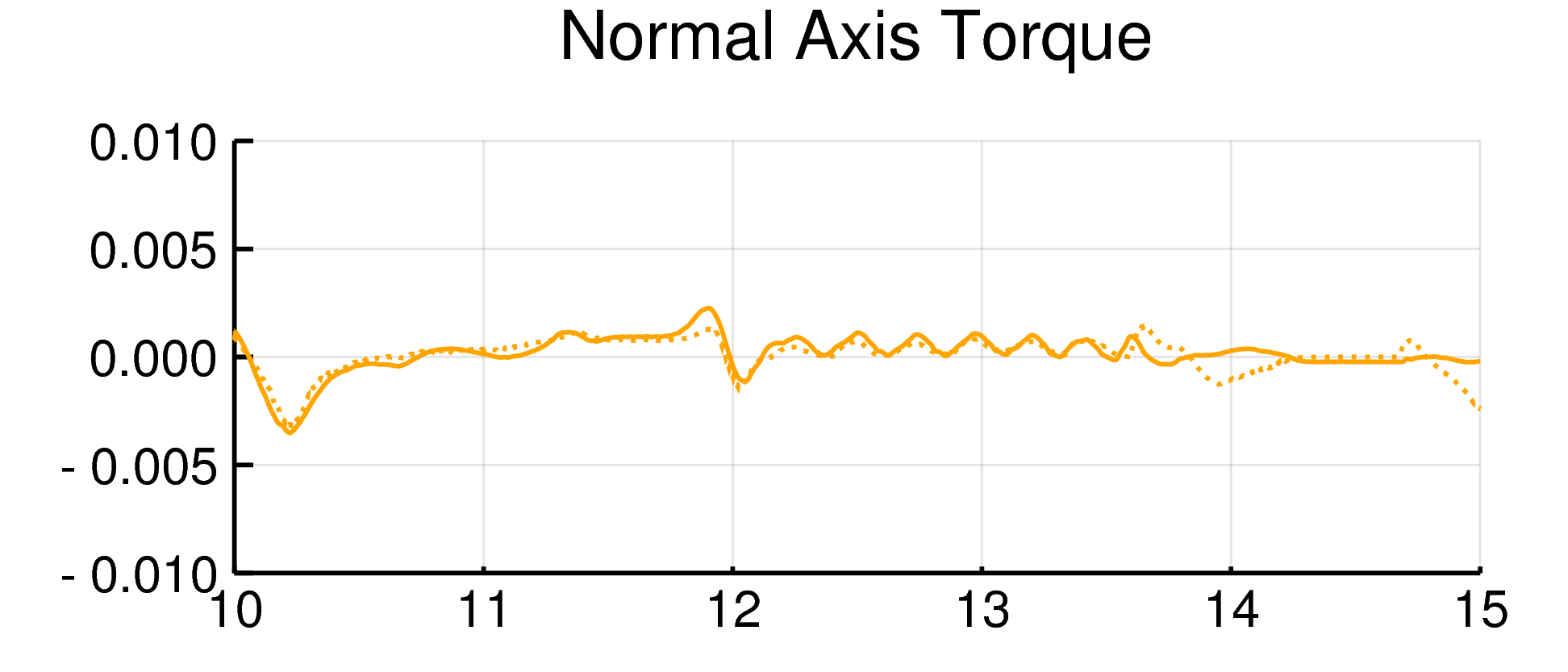}
	\caption{In these plots, we seed our simulator with the state measured from the hardware system. We use the simulator to calculate the instantaneous forces measured by a simulated force sensor, per time-step of real world data, and compare it to the data collected from hardware. These plots are in the frame of the contact, with the force along the normal axis being greatest. Note that the Y-axis of the Normal Axis Torque plot is different from the other torque plots.}
    \label{fig:forcetorque}
\end{figure*}

The recorded behaviors are represented as a list of sensor readings ${\bf S}=\{s_1,s_2,\ldots,s_n\}$ and motor torques ${\bf U}=\{u_1,u_2,\ldots,u_n\}$. State estimation means finding a trajectory of states ${\bf Q}=\{q_1,q_2,\ldots,q_n\}$. Each state is a vector $q_i=(\theta_1,\ldots,\theta_{k'},x,y,z,q_w,q_x,q_y,q_z)$, representing joint angles and object position. We also perform system ID which is finding the set of parameters $\textbf{P}$, which include coefficients of friction, contact softness, damping coefficients, link inertias and others. We then pose the system ID and estimation joint problems as the following optimization problem: $$\min_{{\bf P},{\bf Q}} \sum_{i=1\ldots n}\|\hat{\tau_i}-u_i \|^{*_1} + \sum_{i=1\ldots n}\|\hat{s_i}-s_i \|^{*_2}$$ where $\hat{\tau_i}$ (predicted control signal) and $\hat{s_i}$ (predicted sensor outputs) are computed by the inverse dynamics generative model of MuJoCo: $(\hat{\tau}_i,\hat{s}_i)=\textrm{mj\_inverse}(q_{i-1},q_i,q_{i+1})$. The optimization problem is solved via Gauss-Newton \cite{kolev2015physically}.

\section{TASK \& EXPERIMENTS}

In this section we first describe the manipulation task used to evaluate learned policy performance, then describe the practical considerations involved in using the NPG algorithm in this work. Finally, we describe two experiments evaluating learned policy performance in both simulation and on hardware.

\subsection{Task Description}

We use the NPG algorithm to learn a pushing task. The goal is to reduce the distance of the object, in this case the cylinder, to a target position as much as possible. This manipulation task requires that contacts can be made and broken multiple times during a pushing movement. As there are no state constraints involved in this RL algorithm, we cannot guarantee that the object will reach the target location (the object can be pushed into an un-reachable location). We feel that this is an acceptable trade-off if we can achieve more robust control over a wider state space.

For these tasks we model the bases of each Phantom as fixed, arrayed roughly equilateral around the object being manipulated--this is to achieve closure around the object. We do not enforce a precise location for the bases to make the manipulation tasks more challenging and expect them to shift during operation regardless.

\subsection{Training Considerations}

Policy training is the process of discovering which actions the controller should take from the current state to achieve high reward. As such, it has implications for how well-performing the final policy is. Training structure informs the policy of good behavior, but is contrasted with the time required to craft the reward function. In this task's case, we use a very simple reward structure. In addition to the primary reward of reducing the distance between the object and the goal location, we provide the reward function with terms to reduce the distance between each finger tip and the object. This kind of hint term is common in both reinforcement learnin g and trajectory optimization. There is also a control cost, $a$, that penalizes using too much torque. The entire reward function at time $t$ is as follows:

\begin{equation}
R_t(s,a) = 1 - 3\|O_{xy} - G_{xy}\| \\
- \sum_{i=1}^{3}{\|f_i - O_{xy}\|} \\
- 0.1a^2
\end{equation}

Where $O_{xy}$ is the current position on the $xy$-plane of the object, $G_{xy}$ is the goal position, and $f_i$ is the Phantom end effector position. The state $s$ consists of joint positions, velocities, and goal position. The actions ($a$) are torques.

The initial state of each trajectory rollout is with the Phantom robots at randomized joint positions, deliberately not contacting the cylinder object. The cylinder location is kept at the origin on the $xy$-plane, but the desired goal location is set to a uniformly random point within a 12cm diameter circle around the origin. To have more diverse initializations and to encourage robust policies, new initial states have a chance of starting at some state from one of the previous iteration's trajectories, provided the previous trajectory had a high reward. If the initial state was a continuation of the previous trajectory, the target location was again randomized: performing well previously only gives an initial state, and the policy needs to learn to push the object to a new goal. This is similar to a procedure outlined in \cite{mordatch2015interactive} for training interactive and robust policies.

Finally, to further encourage robust behavior, we vary the location of the base of the Phantom robots by adding Gaussian noise before each rollout. The standard deviation of this noise is 0.5cm. In this way an ensemble of models is used in discovering robust behavior. As discussed above, we expect to have imprecisely measured their base locations and for each base to potentially shift during operation. We examine this effect more closely as one of our experiments.

\begin{figure*}[t!]
\begin{center}
\vspace*{3mm}
\includegraphics[width=0.98\textwidth]{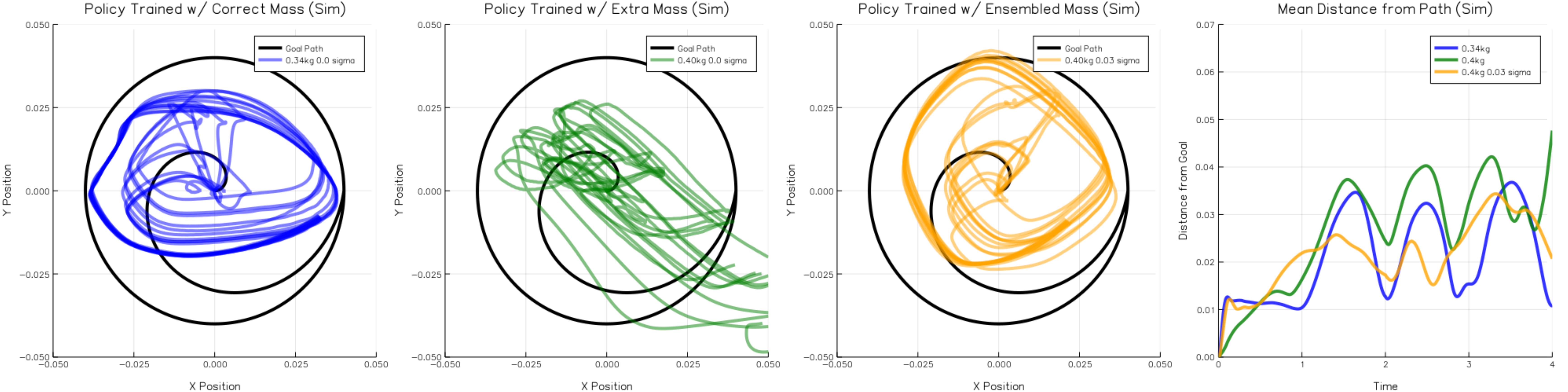}
\includegraphics[width=0.98\textwidth]{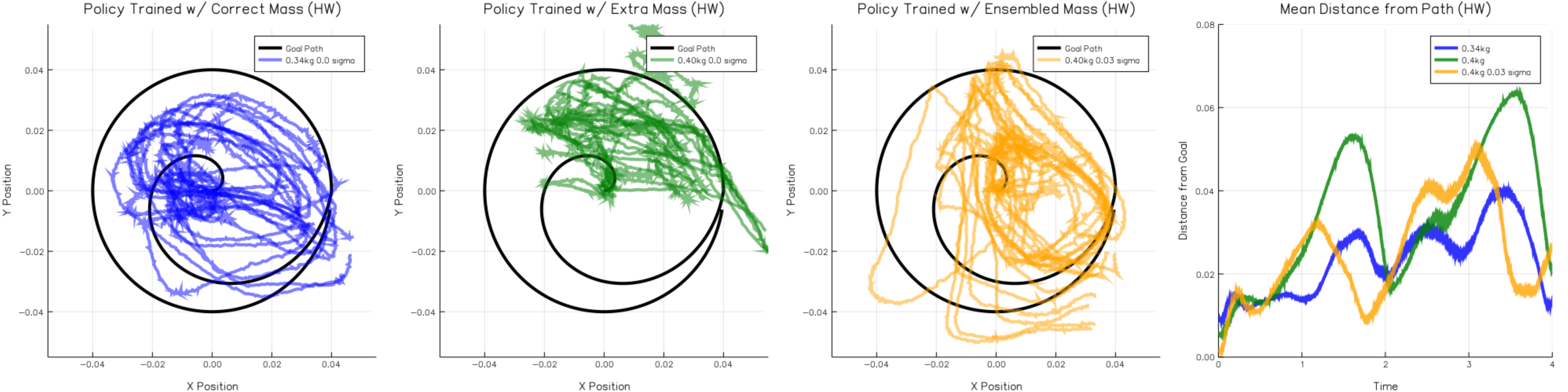}
\end{center}
\caption{10 rollouts are performed where the target position of the object is the path that spirals from the center outward (in black) and then performs a full circular sweep (the plots represent a top-down view). We compare three differently trained policies: one where the mass of the object cylinder is 0.34kg, one where the mass is increased by 20 percent (to 0.4kg), and finally, we train a policy with the incorrect mass, but add model noise (at standard deviation 0.03) during training to create an ensemble. We evaluate these policies on the correct (0.34kg) mass in both simulation and on hardware. In both, the policy trained with the incorrect mass does not track the goal path, sometimes even losing the object. We also calculate the per time-step error from the goal path, averaged from all 10 rollouts (right-most plots); there is usually a non-zero error between the object and the reference due to the feedback policy having to 'catch up' to the reference.}
\label{fig:ensemble}
\end{figure*}

\begin{figure}[h]
	\centering
    \vspace*{3mm}
	\includegraphics[width=0.48\textwidth]{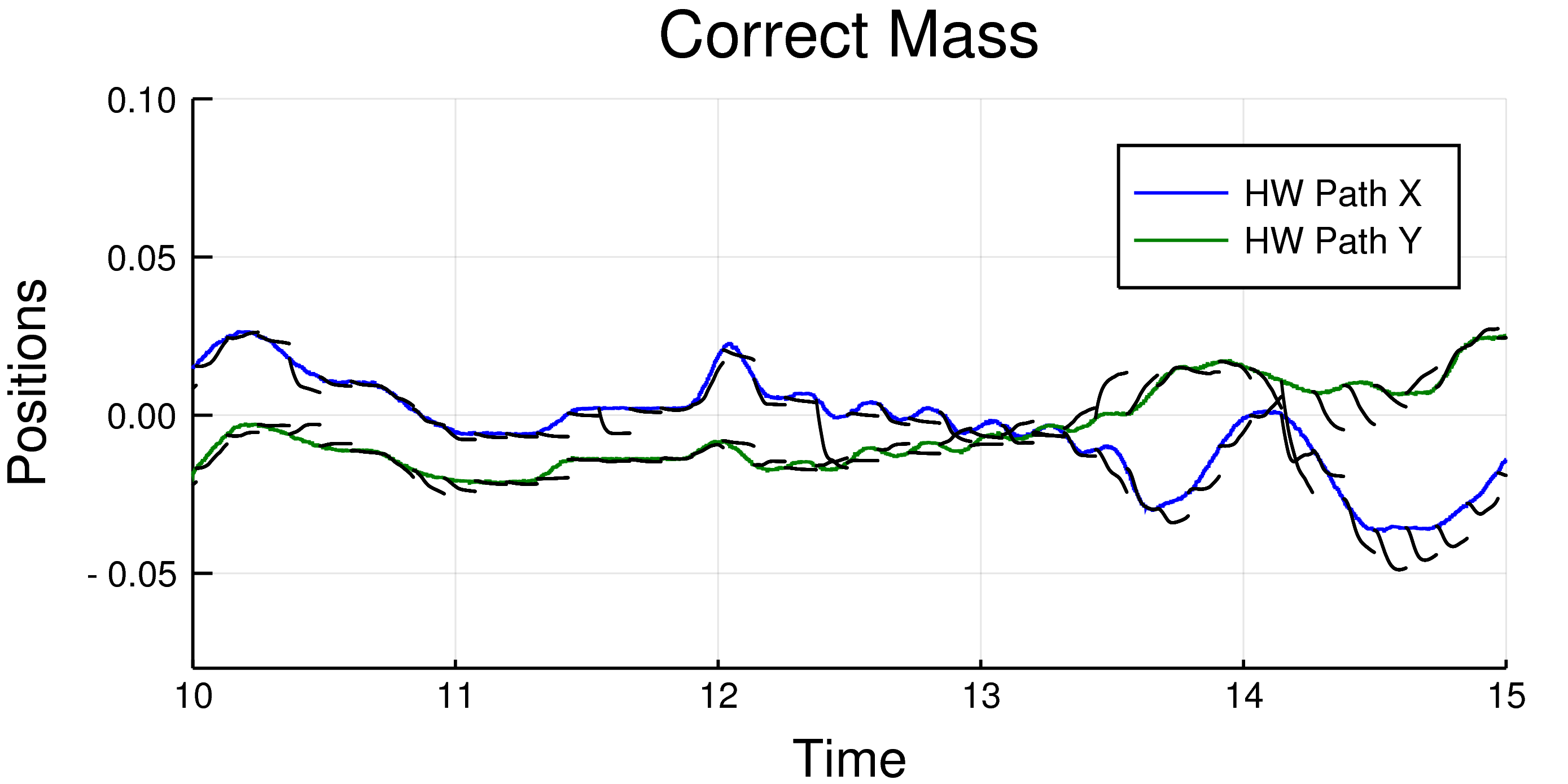}
	\includegraphics[width=0.48\textwidth]{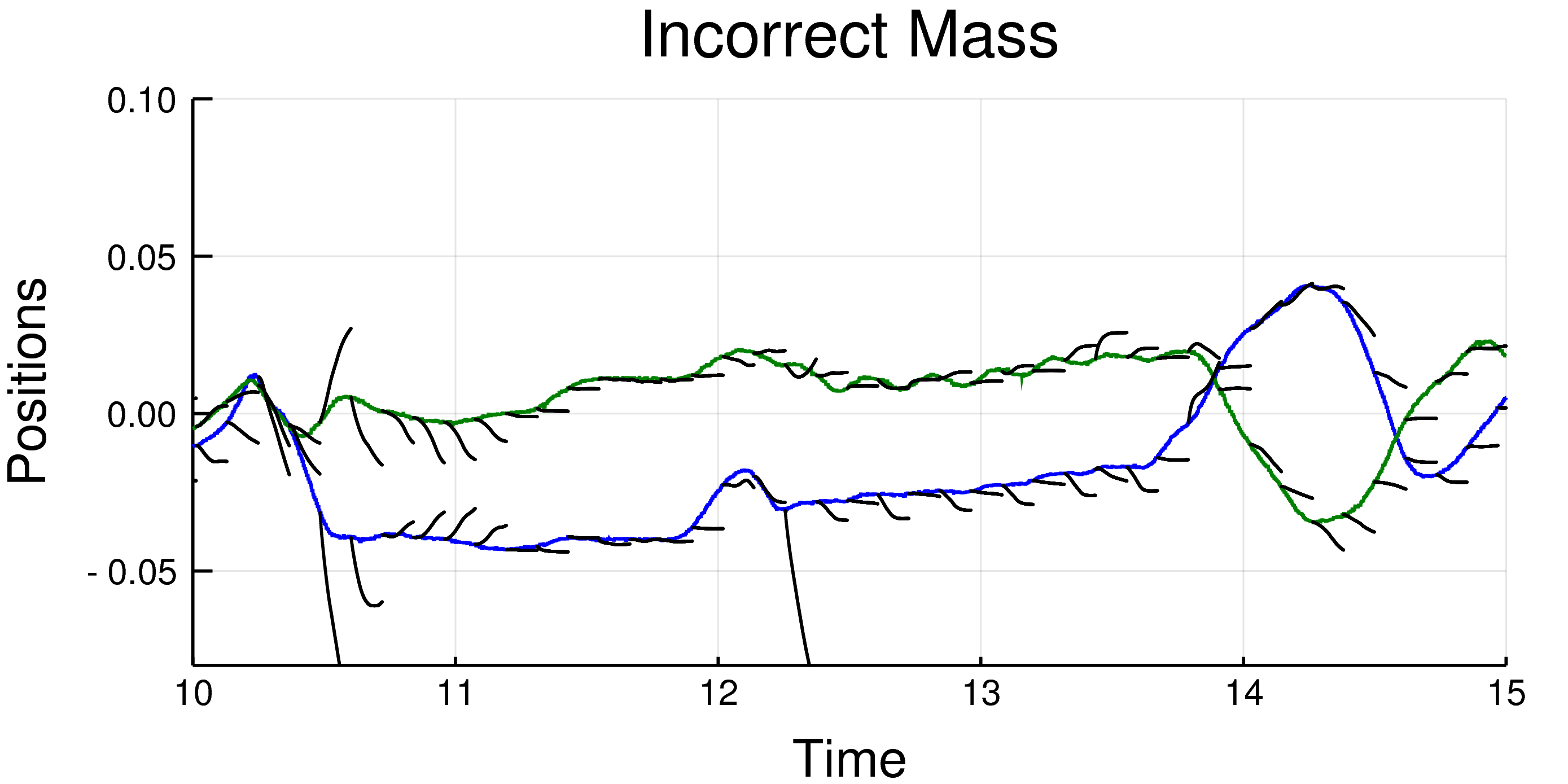}
	\includegraphics[width=0.48\textwidth]{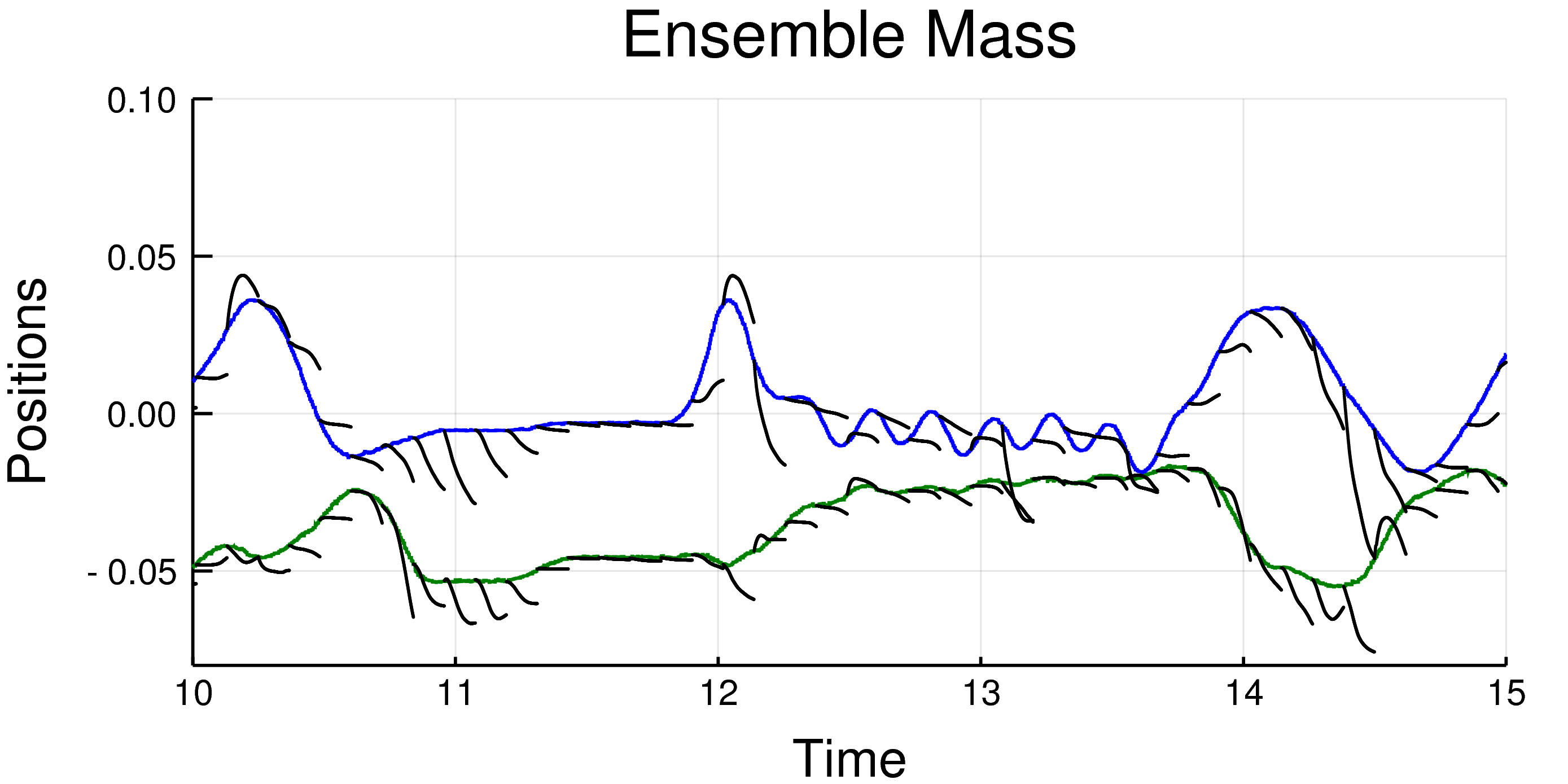}
	\caption{We show the effects of different controllers. We seed our simulator with the hardware's state, and, in simulation, perform a rollout of 200 time-steps -- about 0.1 seconds. This is rendered as the black lines above. The correct policy (trained with measured mass), has rollouts that closely match the measured hardware state data. The incorrect policy (trained with an incorrect mass), performs differently in simulation. The remaining ensemble policy performs better than the incorrect one; this demonstrates that a 'safe' policy can be learned to at least begin an iterative data collection process. While it could be expected that the policies perform similarly in both simulation and hardware, we see that it is not the case here. A policy trained on an incorrect model would over-fit to the incorrect model, and changing to one of two different models (i.e. simulation or hardware) can have un-intuitive effects.}
    \label{fig:simreal}
\end{figure}

\begin{figure}[h]
\begin{center}
\vspace*{2mm}
\includegraphics[width=0.48\textwidth]{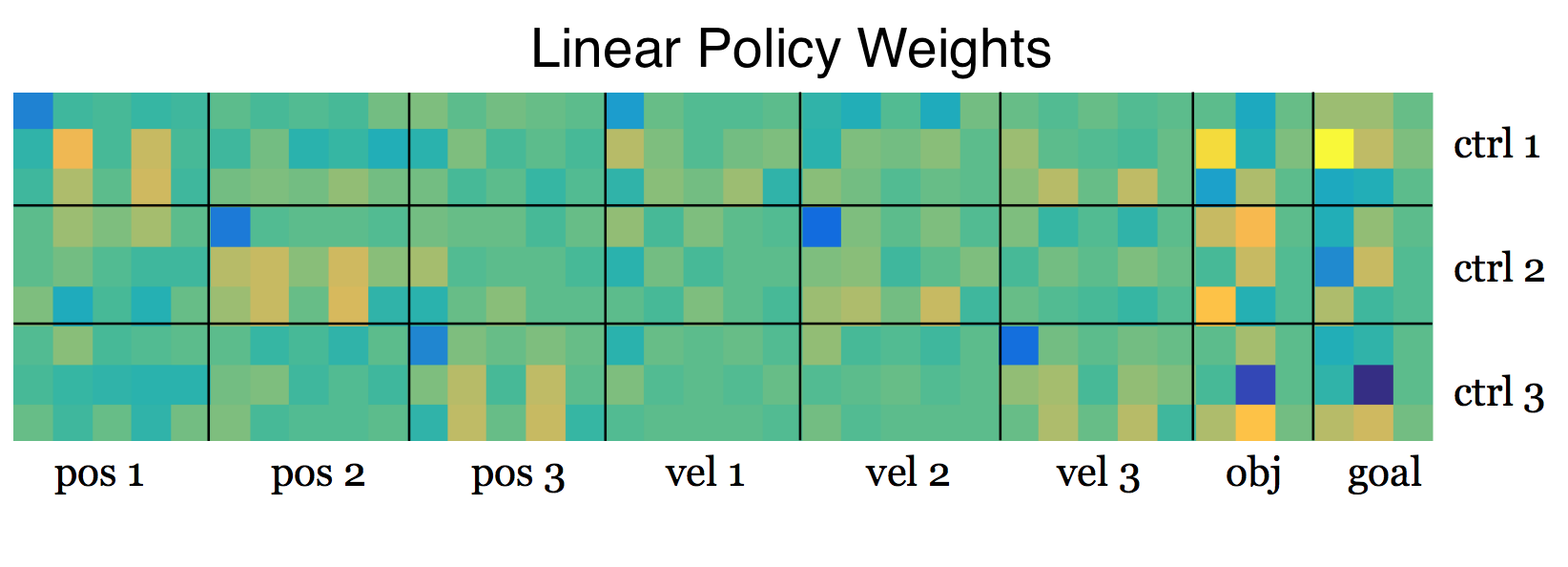}
\end{center}
    \vspace*{-3mm}
\caption{We render the policy parameters. A distinct pattern can be seen in the three negative weights in the top left of the pos and vel blocks. These correspond to the control outputs for the each of the three Phantom's turret joint; as each robot is sensitive, the policy outputs a negative gain to avoid large control forces. Additionally, we can see that the first and second Phantom contributes primarily the object's X location, while the third Phantom handles the Y location. This linear policy can be likened to a hand crafted feedback policy, but the numerous non-zero values would be unintuitive to select for a human. Presumably they can be used to contribute to the controllers robustness as learned through this method.}
\label{fig:heatmap}
\end{figure}

\subsection{Experiments}

We devised two experiments to explore the validity of the NPG reinforcement learning algorithm to discover robust policies for difficult robotic manipulation tasks.

First, we collect runtime data of positions, velocities, and force-torque measurements from a sensor equipped Phantom using the best performing controller we have learned. We use this hardware data (positions and calculated velocities of the system) to seed our simulator, where instantaneous forces are calculated using a simulated force-torque sensor. This data is compared to force-torque data collected from the hardware. Instantaneous force differences highlight the inaccuracies between a model in simulation and data in the real world that eventually lead to divergence.

We also compare short trajectory rollouts in simulation that have been seeded with data collected with hardware. This compares the policy's behavior, not the system's, as we wish to examine the performance of the policy in both hardware and simulation. Ideally, if the simulated environment matches the hardware, it can be taken as an indication that the system identification has been performed correctly. Secondly, we would like to compare the behavior of the learned control policy. From the perspective of task completion, the similarity or divergence of sim and real is less important as long as the robot completes the tasks satisfactorily. Said another way, poor system identification or sim/real divergence matters less if the robot gets the job done.
For these experiments, the target location is set by the user by moving a second tracked object above the cylinder. This data was recorded and used to collect the above datasets for analysis.


As a second experiment direction, we show how the effects of model ensembles during training affect robustness and feasibility. To do this, we explicitly vary the mass of the object being manipulated. The object (cylinder) was measured to be 0.34kg in mass, therefore, we train a policy with the mass set to this value. The object's mass was chosen to be modified due to the very visible effects an incorrect mass would have on performance. We train two additional policies, both with a mass of 0.4kg (approximately 20\% more mass). One of the additional policies is trained with an approximated ensemble: we add Gaussian noise to the object mass parameter with standard deviation of 0.03 (30 grams). All three policies are evaluated in simulation with a correctly measured object mass, and in the real world with our 0.34kg cylinder.

To evaluate the policies, we calculate a path for the target to follow. The path is a spiral from the origin moving outward until it achieves a radius from the origin of 4 cm, at which it changes to a circular path and makes a full rotation, still at a radius of 4cm. This takes 4 seconds to complete. This path was programmatically set in both the simulator and on the real hardware to be consistent. This object ideally follows this trajectory path, as it presents a very visible means to explore policy performance.

\begin{table}[h]
	\caption{Average Distance from Target, 10 Rollouts}
	\label{table_example}
	\begin{center}
		\begin{tabular}{l r r}
			& Sim & Hardware \\
			\hline
			0.34kg Policy (correct)  & 2.1cm  & 2.33cm \\
			0.4kg Policy (incorrect) & 2.65cm & 3.4cm \\
			0.4kg Policy ensemble    & 2.15cm & 2.52cm \\
		\end{tabular}
	\end{center}
\end{table}

\section{Results}

The results for the two experiments are presented as follows, with additional discussion in the next section.

\subsection{Simulation vs Hardware}

We show comparisons between calculated forces and torques in simulation and hardware in figure \ref{fig:forcetorque}. Our simulated values closely match the sensed hardware values. However, the discretization of hardware sensors for the joint positions and velocities are not as precise as simulation, which may result in a different calculation of instantaneous forces. While MuJoCo can represent soft contacts, the parameters defining them were not identified accurately. Critically, we can see that when contact is not being made, the sensors, simulated and hardware, are in agreement.

We find that our learned controllers are still able to perform well at task completion, despite differences between simulation and hardware. We can see in figure \ref{fig:simreal} that for the correct policy (learned with a correct model), when we perform a rollout in simulation based on hardware data, the simulated rollout is close to the data collected from hardware. The policy performance in simulation is close to the policy performance in hardware. This is not the case for the policy trained using incorrect mass and the ensemble policy, where the simulated rollout is different from the hardware data. We hypothesize that policies trained on specific models over-fit to these models, and take advantage of the specifics of the model to complete the task. 
Despite the correct simulation being similar to hardware, the controller's behavior could cause divergence on whatever remaining small parameter differences. The ensemble policy, as expected, lies somewhere between the correct and incorrect policy (trained with incorrect parameters).

\subsection{Training with Ensembles}

We find that training policies with model ensembles to be particularly helpful. Despite being given a very incorrect mass of the object, the policy trained with the ensemble performed very well (figure \ref{fig:ensemble}). In addition to performing well in simulation, we found it to perform nearly as well as the correctly trained policy on hardware. Given that these are learned feedback controllers, there is a distinct lag of the object behind the desired reference trajectory. Table 1 quantifies the error from the reference trajectory across the whole length of the action. This mirrors our comparison in the previous section, where the correct policy performs comparably in both hardware and simulation, with the other policies less so. This is important to note given the poor performance of the incorrect policy: this task's training is indeed sensitive to this model parameter. The implication of the ensemble approach is not just that it can overcome poor or incorrect modeling, but can provide a safe initial policy to collect valuable data to improve the model.

\section{DISCUSSION}

Our results suggest two interesting observations. 
Simulation can provide a safe backdrop to test and develop non-intuitive controllers (see figure \ref{fig:heatmap}). This controller was developed for a robotic system without an intermediate controller such as PID, and without human demonstrations to shape the behavior. We also eschewed the use of a state estimator during training and run-time as this would add additional modeling reliance and complexity. Our simulation based policy learning approach also conveniently allows for building robust controllers by creating ensembles of models by varying physical parameters.

We show how simulated ensemble methods provide two major benefits. Firstly, it can partially make up for incorrectly measured / identified model parameters. This benefit should be obvious: it can be difficult to measure model parameters affecting nonlinear physical phenomena. Additionally, training in an ensemble has the added benefit of allowing for more conservative policies to enable appropriate data collection for actual model improvement. A natural extension of this observation would be full model adaptation using a technique such as EPopt \cite{rajeswaran2016epopt}.

Model adaptation provides a bridge between model-based and model-free methods. Leveraging a model in simulation can provide a useful policy to begin robot operation, which can subsequently be fine-tuned on hardware in a model-free mode. Very dynamic behaviors may not be suited to direct hardware training without significant human imposed safety constraints, which may take significant time to develop and may not account for all use cases. Given that most robots are manufactured using modern techniques, a model to be used in simulation is very likely to exist, and this model should be leveraged to obtain better policies.



\section*{Acknowledgements}
This work was supported in part by the NSF. The authors would like to thank Vikash Kumar and the anonymous reviewers for valuable feedback.

\clearpage
\bibliographystyle{IEEEtran}
\bibliography{references}

\end{document}